\documentclass[journal]{IEEEtran}
\usepackage[utf8]{inputenc}

\usepackage[flushleft]{threeparttable}
\usepackage{amsmath}
\usepackage{amssymb}
\usepackage{multirow}
\usepackage{amsfonts}
\usepackage{graphicx}
\usepackage{subfig}
\usepackage{booktabs}
\usepackage{afterpage}
\usepackage{hyperref}
\usepackage{makecell}
\usepackage{algpseudocode, algorithm, algorithmicx}

\graphicspath{{images/},{TIFS_final_submission/authors/}}

\begin{document}



\title{Meta-learning for fast classifier adaptation to new users of Signature Verification systems}

\author{Luiz~G.~Hafemann,
	Robert Sabourin,~\IEEEmembership{Member,~IEEE,}
	and~Luiz~S.~Oliveira
	\thanks{L. G. Hafemann and R. Sabourin are with the Laboratoire d'imagerie, de vision et d'intelligence artificielle, \'Ecole de technologie sup\'erieure, Universit\'e du Qu\'ebec, Montreal, Canada. (e-mail: luiz.gh@mailbox.org, robert.sabourin@etsmtl.ca)}
	\thanks{L. S. Oliveira is with the Department of Informatics, Federal University of Parana, Curitiba, Brazil (e-mail:lesoliveira@inf.ufpr.br)}.
	\thanks{This work was supported by the Fonds de recherche du Qu\'ebec - Nature et technologies (FRQNT), the CNPq grant \#206318/2014-6 and by the grant RGPIN-2015-04490 to Robert Sabourin from the NSERC of Canada.}
}

\IEEEpubid{1556-6013 \copyright 2019 IEEE. Personal use is permitted, but republication/redistribution  requires  IEEE permission. }

\maketitle

\begin{abstract}
	
Offline Handwritten Signature verification presents a challenging Pattern Recognition problem, where only knowledge of the positive class is available for training. While classifiers have access to a few genuine signatures for training, during generalization they also need to discriminate forgeries. This is particularly challenging for \emph{skilled} forgeries, where a forger practices imitating the user's signature, and often is able to create forgeries visually close to the original signatures. Most work in the literature address this issue by training for a surrogate objective: discriminating genuine signatures of a user and random forgeries (signatures from other users).
In this work, we propose a solution for this problem based on meta-learning, where there are two levels of learning: a task-level (where a task is to learn a classifier for a given user) and a meta-level (learning across tasks). In particular, the meta-learner guides the adaptation (learning) of a classifier for each user, which is a lightweight operation that only requires genuine signatures. The meta-learning procedure learns what is common for the classification across different users. In a scenario where skilled forgeries from a subset of users are available, the meta-learner can guide classifiers to be discriminative of skilled forgeries even if the classifiers themselves do not use skilled forgeries for learning.
Experiments conducted on the GPDS-960 dataset show improved performance compared to Writer-Independent systems, and achieve results comparable to state-of-the-art Writer-Dependent systems in the regime of few samples per user (5 reference signatures). 

\end{abstract}


\section{Introduction}

Offline Handwritten signature verification remains a challenging problem in the presence of skilled forgeries, where the forger has access to the user's signature and practices imitating it \cite{hafemann_offline_2017}. This problem is particularly challenging since in a practical application scenario we cannot expect to have access to skilled forgeries for every user in the system for training the classifiers.

This problem is mainly addressed in three ways in the literature: (i) training a classifier for each user using a surrogate objective, where the negative samples are genuine signatures from other users (called \emph{random forgeries} in this context) \cite{vargas_off-line_2010, yilmaz_score_2016, hafemann_pr_2017} (ii) training a one-class classifier for each user \cite{guerbai_effective_2015}; (iii) training a global, writer-independent classifier \cite{kumar_writer-independent_2012, eskander_hybrid_2013, rantzsch_signature_2016}. The first alternative (Writer Dependent (WD) classification) optimizes a surrogate objective, which therefore can be sub-optimal. The second alternative (one class Writer Dependent classification) is an appropriate formulation of the problem, but empirical results show that this approach performs worse than the first. A possible reason is that for signature verification tasks we normally have only a small number of samples per user, which makes it hard to estimate the support (or probability density) of the positive class. For instance, recent work considers a feature space in $\mathbb{R}^{2048}$, while the number of signatures from one individual can be as low as 1-5 in practical applications  \cite{hafemann_offline_2017, hafemann_pr_2017}.
Lastly, the third alternative (Writer Independent (WI) classification) alleviates the problem of a small number of samples per user by transforming the problem in a binary classification problem: comparing a query signature with a reference (template) signature, where the same classifier is used for all users \cite{ rivard_multi-feature_2013, eskander_hybrid_2013}. However, empirically these approaches also show worse performance than WD classification, at least when the number of signatures available for training (per user) is larger than 1 \cite{hafemann_pr_2017}. We hypothesize that a reason for this gap in performance is that the WI classifiers compare a query signature with a reference signature one at a time, while the WD classifiers are trained with multiple references at the same time, and therefore can better estimate the invariances in a person's signature (intra-class variation).

Considering different approaches, WD classification (alternative (i) above) shows better empirical performance \cite{hafemann_offline_2017}. However, this approach has other shortcomings compared to WI approaches: they require training a classifier for each user, which is not desirable in some scenarios: For instance, when the number of users is very large, and each user do not use the system often - many classifiers are trained but are almost never used. Also, in the cases where features are learned from data (e.g. \cite{hafemann_pr_2017}), if we want to change the feature representation, for instance by training with new data, it is not straightforward to incorporate the new features without re-training all WD classifiers in the system, while a global (WI) classifier would not require any extra step. WI systems also naturally handles the issue of adding more signatures to the reference set. 

In this work, we propose to formulate the task as a meta-learning problem, inspired by the work of a Forensics Handwritten Expert: the expert acquires knowledge examining genuine signatures and forgeries from several people along his/her training and work experience. For a new case, along with knowledge of signatures from the individual, this previous experience is also used when analyzing a signature of interest. 

The main contributions of this paper are:
\begin{itemize}
	\IEEEpubidadjcol
	\item We formulate the signature verification task as a meta-learning problem, considering a meta-learner that learns across-tasks (classification for specific individuals), that is subsequently adapted to a particular user in order to make a prediction on a query signature.
	\item We extend Model Agnostic Meta Learning (MAML) \cite{finn_model-agnostic_2017}, to consider different loss functions during classifier adaptation and meta-learning, to address the issue of partial-knowledge during training. 
	\item The resulting system is as scalable as a WI system (there is a single meta-model), but that is also adaptable for individual users with a lightweight operation (a few gradient descent steps). 	Additionally, contrary to other work that learns representations to train WD classifiers (\cite{hafemann_pr_2017}), not only the final classification layer is adapted to the new user, but the \emph{feature representation} is also adapted.  
	\item We evaluate the approach in four widely used datasets, achieving results comparable to state-of-the-art on the GPDS-960 dataset. Finally, we discuss the limitations of the approach, most notably the requirement of using data from a large number of users for training, and worse results when transferring the meta-learner to the other datasets. Code to reproduce the experiments can be found at \url{https://github.com/luizgh/sigver}.
\end{itemize}

The paper is organized as follows: section \ref{sec:related} reviews the related work on signature verification and meta-learning. Section  \ref{sec:proposed} introduces the formulation of signature verification as a meta-learning problem, and the proposed algorithm. Section \ref{sec:experimental_protocol} describes the experimental protocol, and section \ref{sec:results} presents and discusses our results. Finally, section \ref{sec:conclusion} concludes the paper.


\section{Related Work}
\label{sec:related}

The objective of signature verification systems is to classify a query signature as being genuine (produced by the claimed individual), or a forgery (produced by another person). In the Pattern Recognition community, different forgeries are considered: Random forgeries - in which the forger has no knowledge of the user's signature, and use his signature instead; Simple forgeries - in which the forger knows the person's name, but not their signature; Skilled forgeries - where the forger has access to the user's signature, and practices imitating it. While the problem of distinguishing random and simple forgeries is relatively easy (i.e. low error rates in state-of-the-art classifiers), skilled forgeries still present a significant challenge for classification.

These systems can be broadly categorized as Writer-Dependent (WD, also called User-Dependent) and Writer-Independent (WI, also called User-Independent). For Writer-Dependent classifiers, we consider a dataset for each user $\{x, y\}_{i=1}^n$, where $x$ are signatures, and $y$ indicate whether they are genuine signatures from the user ($y=1$) or random forgeries ($y=0$) \cite{vargas_off-line_2010, yilmaz_score_2016, hafemann_pr_2017}. Some work consider one-class WD classifiers, in which only genuine signatures from the user are used for training (only $y=1$) \cite{guerbai_effective_2015}. For WI classifiers, there are two main approaches: training a single classifier in a \emph{dissimilarity} space, and \emph{metric learning} approaches. In the first case, the training samples are difference of feature vectors: $| \phi(x_1) - \phi(x_2) |$, with $y=1$ if both signatures are from the same user, and $y=0$ otherwise \cite{kumar_writer-independent_2012, eskander_hybrid_2013}. The metric learning approaches use a siamese network architecture \cite{bromley_signature_1994}, which takes two signatures $(x_1, x_2)$ as input, and outputs a metric (distance) between them. 

\begin{figure}
	\centering
	\includegraphics[scale=0.5]{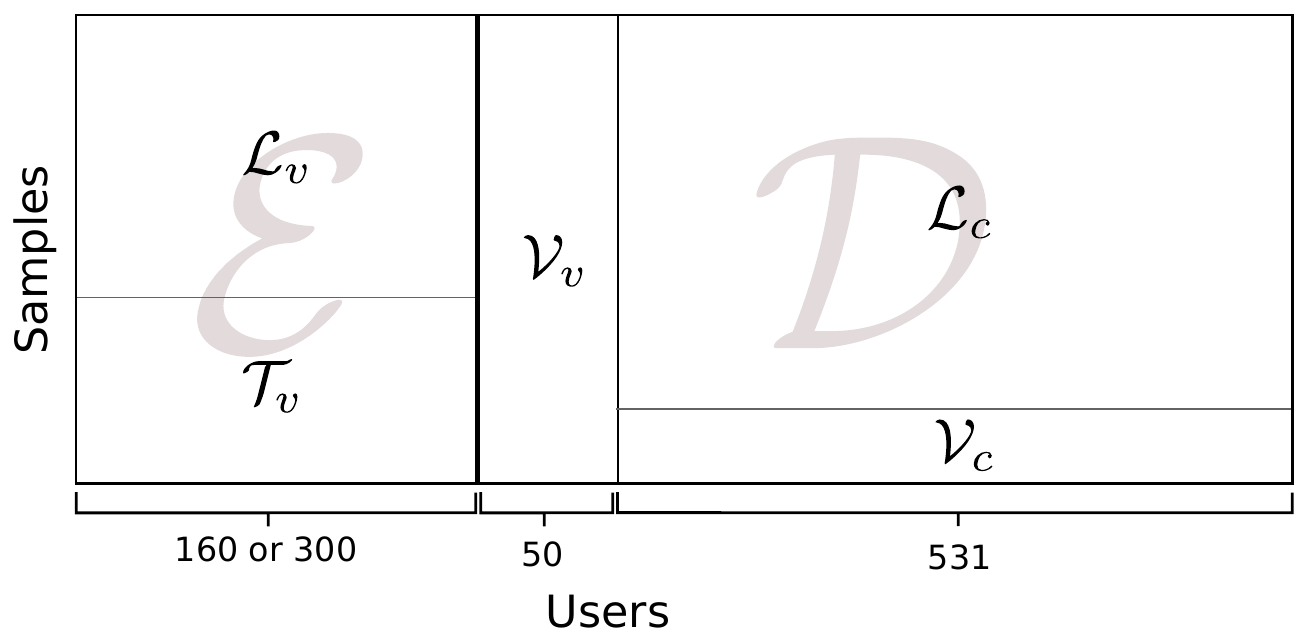}
	\caption{Common dataset separation for Feature Learning followed by WD classification, on the GPDS dataset. Features are learned in $\mathcal{D}$. Model selection is conducted in $\mathcal{V}_v$. The system is evaluated by training WD classifiers for the exploitation set $\mathcal{E}$. \cite{hafemann_pr_2017}}
	\label{fig:dataset_separation}
\end{figure}

Recent work on signature verification rely on feature learning methods \cite{hafemann_ijcnn_2016, hafemann_pr_2017, hafemann2018fixed, zois_hierarchical_2017, rantzsch_signature_2016}, in which learning is conducted directly from signature pixels, instead of relying on handcrafted feature extractors. In this case, a function $\phi(x)$ is learned to \emph{extract features} from signature images $x$, by training using a surrogate objective, e.g. dictionary learning \cite{zois_parsimonious_2017, zois_hierarchical_2017}, or classifying the user that produce the signatures \cite{hafemann_pr_2017}. For instance, the SigNet model \cite{hafemann_pr_2017} is a Convolutional Neural Network trained with the following objective:

\begin{equation}
L = -\sum_j{y_{ij} \log{P(y_{j} | X_i})}
\end{equation}
Where $X_i$ is a signature and $y_i$ is the user that wrote the signature. Therefore, the network is trained to obtain a representation space where signatures from different people are linearly separable \cite{hafemann_pr_2017}. This feature representation is learned from a Development dataset $\mathcal{D}$, which is then used to extract features and train Writer-Dependent classifiers for a disjoint set of users (exploitation set $\mathcal{E}$) - a diagram of dataset separation is shown in Figure \ref{fig:dataset_separation}. 
While this approach achieved state-of-the-art verification performance, we note that the feature learning process does not directly optimize for the final objective of the system, which is distinguish genuine signatures and forgeries. This is addressed to some extent in the SigNet-F model, by also classifying whether or not the signature is a forgery. However, in that case, the neuron classifying forgeries does not use a reference signature from the user. While this was shown to be helpful in obtaining a good feature representation, this neuron did not generalize in classifying forgeries for unseen users \cite{hafemann_pr_2017}. 

Such methods using feature learning followed by WD classification also have other shortcomings: they require training one classifier for each user, which may be an expensive operation (e.g. best results in \cite{hafemann_ijcnn_2016, hafemann_pr_2017} were reported with an SVM trained with the RBF kernel for each user). If the feature extractor is updated (e.g. trained with more data), then all classifiers need to be retrained. Also, these systems use a fixed representation for all users, and it is possible that adapting the representation for each user would yield improvements in classification performance.


It is also worth noting that, for WI classification, signature verification systems can be trained jointly (feature extraction and classification) \cite{rantzsch_signature_2016}. Despite being jointly trained, such WI systems still perform worse than WD classifiers trained with features learned with surrogate objectives, at least when more than one signature references are used \cite{hafemann_pr_2017}. 
A possible reason for this gap is the fact that WI systems compare the query signature to each reference individually (or comparing with the centroid of the signatures), which is less powerful than training a classifier for the user, in capturing the invariances of the person's signature.

\subsection{Meta-learning}

In a broad sense, meta-learning is concerned with the problem of \emph{learning to learn}, with origins in the 80's and 90's \cite{schmidhuber_1987}, \cite{bengio_learning_1991}. More recently, algorithms based on meta-learning have achieved state-of-the-art results in tasks such as hyperparameter optimization \cite{maclaurin_gradient_2015}, neural network architecture search \cite{baker_designing_2017}), and few-shot learning \cite{ravi_optimization_2016, finn_model-agnostic_2017}. Few-shot learning considers a scenario where only a few samples from each class are available for training, which is similar to actual application scenarios in handwritten signature verification. 


The goal of these meta-learning approaches for few-shot learning is to train a model that can quickly (i.e. in a few iterations) adapt to a new task using only a few samples. A new task in this context refers, for instance, to classify a new object, for which only a few samples are known. Ravi and Larochelle \cite{ravi_optimization_2016} proposed learning an \emph{optimizer} and \emph{initialization} for the tasks (Meta Nets). They propose using a Long short-term memory (LSTM) model to learn the update rule for adapting the network parameters to a new task. Finn et al \cite{finn_model-agnostic_2017} proposed a Model Agnostic Meta Learning (MAML) procedure that does not require any extra parameters. This model optimizes the \emph{sensitivity} of the weights, that is, obtain a feature representation that is highly adaptive, such that a single (or a few) gradient descent iterations are sufficient to optimize to new tasks.



\section{Proposed Method}
\label{sec:proposed}

\begin{figure}
	\centering
	\includegraphics[width=\columnwidth]{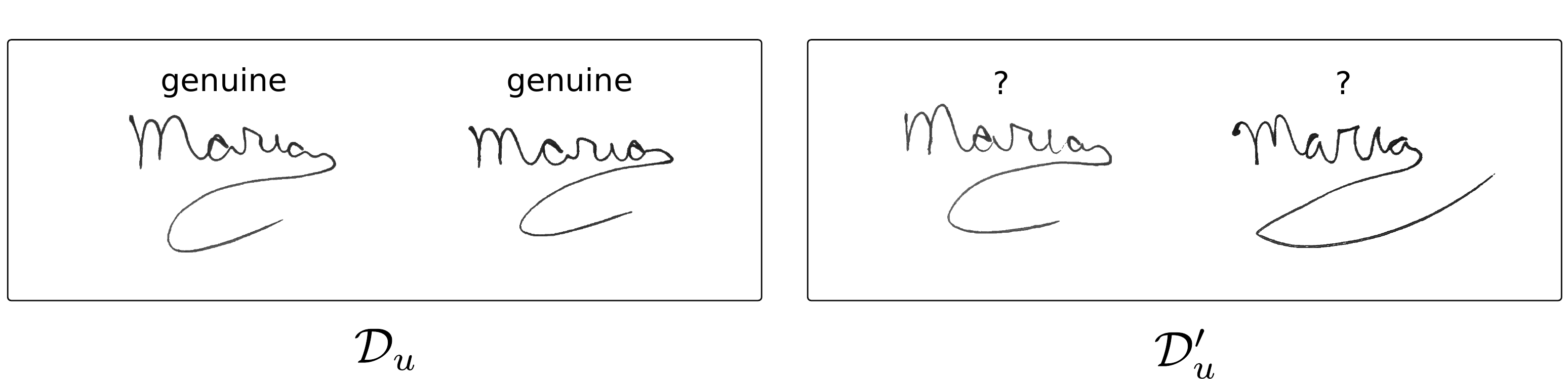}
	\caption{Illustration of the data available for one task (user). Left: the reference (support) set. Right: query samples.}
	\label{fig:data_generalization}
\end{figure}

In this work we propose a meta-learning approach for signature verification. This formulation considers a meta-learner that guides the adaptation of a classifier for each user. We consider that each user describes a \emph{task}: discriminating between genuine signatures (created by the user) and forgeries. Figure \ref{fig:data_generalization} illustrates the data available for one task: we consider a reference (support) dataset that is used for training a classifier that can classify new queries as genuine or forgery.

\begin{figure}
	\centering
	\includegraphics[width=\columnwidth]{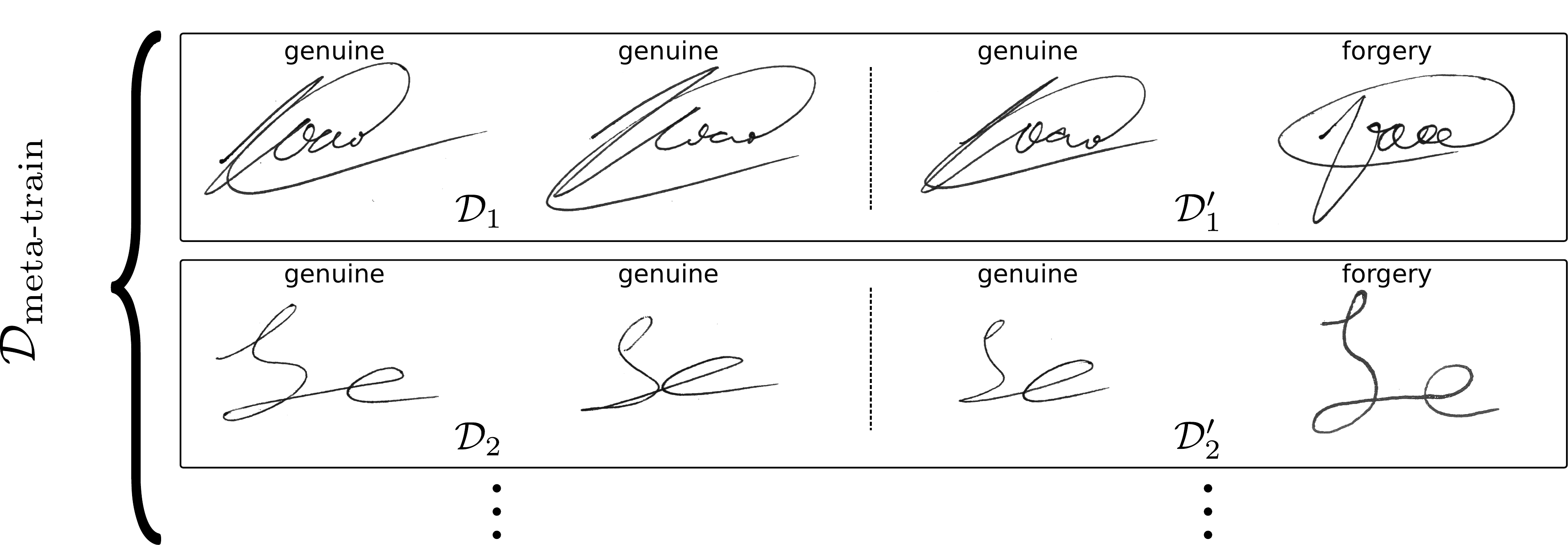}
	\caption{Example of the meta-learning setup. Each user represents an \emph{episode}, where $\mathcal{D}_u$ is used for classifier adaptation and $\mathcal{D}^\prime_u$ is used for meta-update.}
	\label{fig:data_metalearning}
\end{figure}

In a meta-learning setting, we consider that training a classifier for a particular user is guided by a meta-learner, that leverages data from multiple tasks for learning. For this we consider a dataset $\mathcal{D}_\text{meta-train}$, and then evaluate the generalization performance on unseen users $\mathcal{D}_\text{meta-test}$. 

We note that this approach has a direct correspondence to previous work that used feature learning followed by WD classification (section \ref{sec:related}), and here we make the association between the terminology in the meta-learning research and previous work on Signature Verification. In both cases we use a separate set of users for feature learning ($\mathcal{D}_\text{meta-train}$ is analogous to the development set in section \ref{sec:related}), which is then used for to train and test classifiers on a new set of users ($\mathcal{D}_\text{meta-test}$ is analogous to the exploitation set). 
The key differences of meta-learning is that: (i) The loss optimized for feature learning is directly related to the final objective (separate genuine signatures and forgeries); (ii) training a classifier for a new user is a lightweight process (a few gradient descent iterations); (iii) not only the classifier, but the features are also adapted for each user. 

In the next section we formalize the problem of signature verification as a meta-learning task.

\subsection{Problem formulation}

\begin{table}
	\centering
	\caption{Table of symbols}
	\label{tbl:symbols}
	\begin{tabular}{ll}
		\toprule  
		$\mathcal{T}$ & Distribution of tasks (i.e. users) \\
		$\mathcal{T}_u$ & Task for user $u$ \\
		$\mathcal{D}_\text{meta-train}$ & Training set for the meta-learner \\
		$\mathcal{D}_\text{meta-test}$ & Testing set for the meta-learner \\
		$\mathcal{D}_u$ & Samples for weight adaptation for user $u$  \\
		$\mathcal{D}^\prime_u$ & Samples for meta-update for user $u$  \\
		$G_u$ & Genuine signatures for user $u$ \\
		$S_u$ & Skilled forgeries for user $u$ \\
		$\theta$ & Network parameters \\
		$\theta^{(u)}_k$ & Parameters adapted to user $u$ after $k$ descent steps\\
		$L$ & Loss function for weight adaptation \\
		$L^\prime$ & Loss function for meta-update \\
		\bottomrule
	\end{tabular}
\end{table}


We consider that each user describes a task $T_u \in \mathcal{T}$, where the task consists in classifying a signature image as genuine (created by the user) or forgery (not created by the user). A collection of users therefore describes a distribution of tasks $\mathcal{T}$, and the aim of the meta-learner is to explore the structure present in this distribution. We consider a dataset $\mathcal{D}_\text{meta-train}$ containing tasks from $\mathcal{T}$, that is used for meta-learning.
For each user we consider a set $\mathcal{D}_u$, that is used to adapt the classifier, and a set $\mathcal{D}^\prime_u$ that is used for updating the meta-learner.  Lastly, to verify the generalization to unseen users, we consider a set $\mathcal{D}_\text{meta-test}$, that contains data from a disjoint set of users ($\mathcal{D}_\text{meta-train} \cap \mathcal{D}_\text{meta-test} = \emptyset$).  Figure \ref{fig:data_metalearning} illustrates the meta-learning setup, and the symbols used in this paper are listed in Table \ref{tbl:symbols} for clarity.

\begin{figure}
	\centering
	\includegraphics[width=\columnwidth]{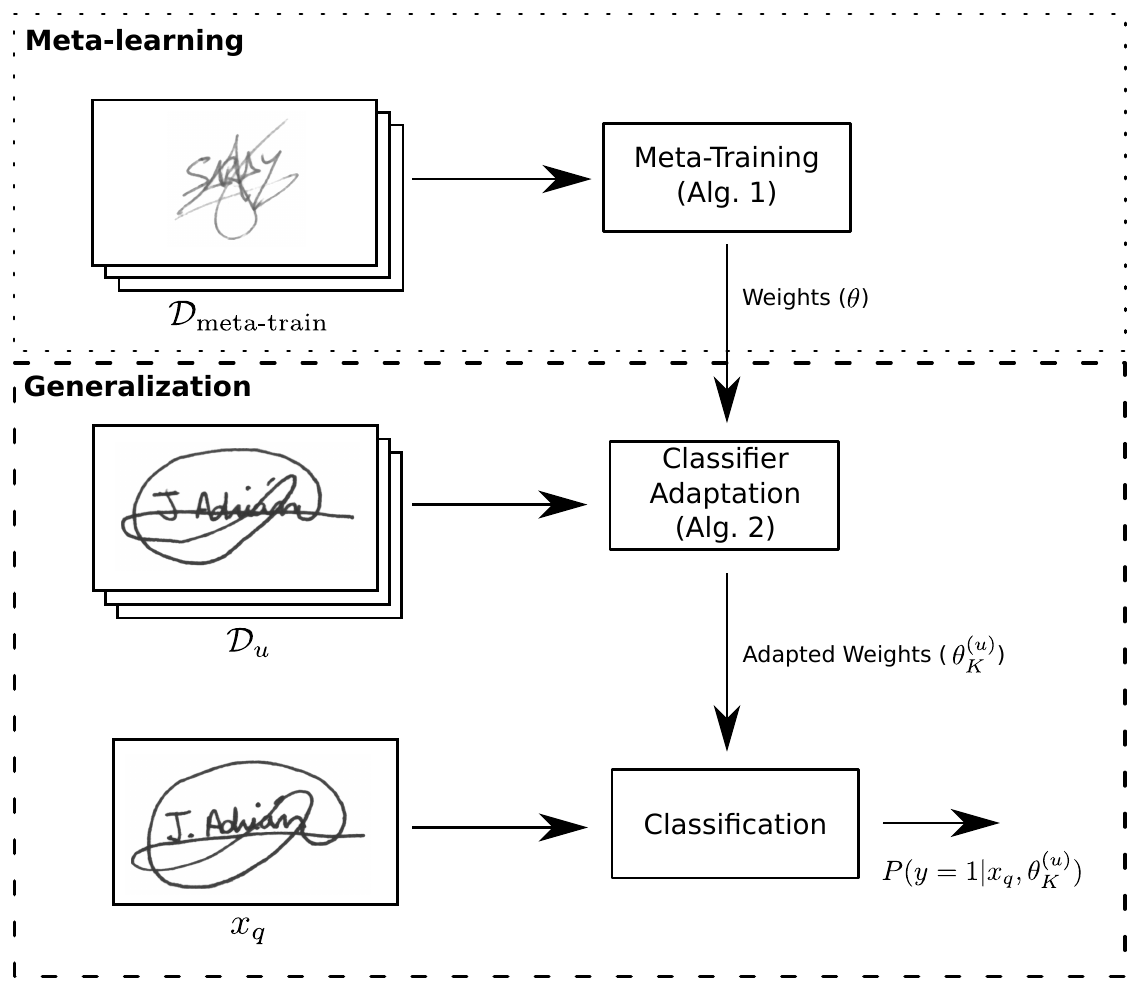}
	\caption{Overview of the meta-learning system for signature verification.}
	\label{fig:architecture}
\end{figure}

\begin{figure*}
	\centering
	\includegraphics[width=\textwidth]{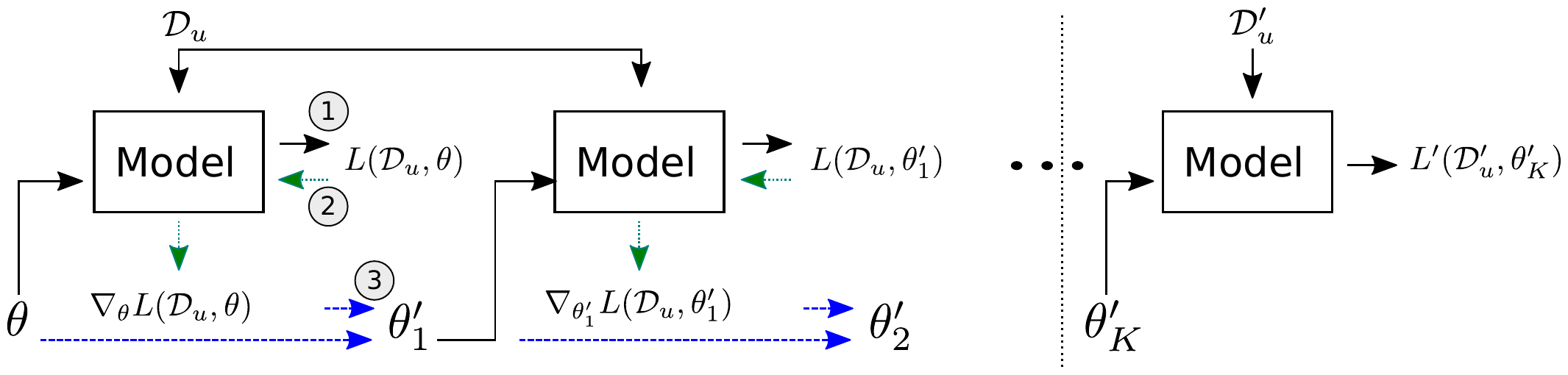}
	\caption{Illustration of one iteration of meta-training for one task $T_u$. Starting with parameters $\theta$, the weights are specialized for the task in $K$ gradient descent steps. Each step involves computing the loss (1), back-propagating the loss w.r.t to $\theta'_{k-1}$ (2) and updating the weights (3). For the meta-update, the loss $L^\prime$ is backpropagated through the entire chain (from $L'$ back to the initial $\theta$), computing $\nabla_\theta L^\prime(\mathcal{D}^\prime_u, \theta^u_K) $. }
	\label{fig:inner_optimization}
\end{figure*}

\subsection{Model Agnostic Meta-Learning for signature verification}
\label{sec:maml_for_sigver}

In this work we propose an extended version of Model-Agnostic Meta-Learning (MAML) \cite{finn_model-agnostic_2017}, by considering different criteria for classifier adaptation and meta-learning.
An overview of the system can be seem in figure \ref{fig:architecture}. We consider a development set for meta-training, that consists in learning the weights $\theta$ of a Convolutional Neural Network, that are highly \emph{adaptable} to new tasks. During generalization, for a user $u$, a reference set $\mathcal{D}_u$ is used to adapt the classifier to this user (using $K$ gradient descent steps) obtaining weights $\theta_K^{(u)}$. This adapted classifier is then used to classify a query image $x_q$, obtaining $P(y=1 | x_q, \theta_K^{(u)})$.

\begin{algorithm}
	\caption{Meta-Training algorithm}
	\label{alg:metalearning}
	\begin{algorithmic}[1]
		\Require $M$: Meta-batch size
		\Require $K$: Number of gradient descent steps
		\Require $\alpha$, $\beta$ Learning rates
		\Ensure $\theta$: Meta-learned weights
		\State Randomly initialize $\theta$
		\While {not done}
		\State Sample a batch of tasks $\{T_u\}_{u=1}^M \sim \mathcal{T}$
		\State $\theta_\text{grad} \gets \vec{0}$ 
		\For {$u \gets 1$ to $M$}
		\State Sample $\mathcal{D}_u$  \Comment{Genuine only}
		\State $\theta^\prime_0 \gets \theta$ 
		\For {$k \gets 1$ to $K$} \Comment{Adapt weights to $u$}
		\State $\theta^\prime_k \gets \theta^\prime_{k-1} - \alpha \nabla_{\theta^\prime_{k-1}}{L(\mathcal{D}_u, \theta^\prime_{k-1})}$ 
		\EndFor
		
		\State Sample $\mathcal{D}'_u$ \Comment{Genuine and forgeries}
		
		\State $\theta_\text{grad} \gets \theta_\text{grad} + \frac{1}{M} \nabla_\theta{L^\prime(\mathcal{D}'_u, \theta^\prime_K)}$ 
		
		\EndFor
		\State $\theta \gets \theta - \beta \theta_\text{grad} $ \Comment{Meta-update}
		\EndWhile
		
	\end{algorithmic}
\end{algorithm}

Algorithm \ref{alg:metalearning} describes the full meta-training algorithm. Meta-training is conducted in \emph{episodes} (Figure \ref{fig:data_metalearning}). In each episode, the classifier is adapted to a particular user using $\mathcal{D}_u$ (lines 7 to 10), and the adapted classifier is used to classify the set $\mathcal{D}'_u$.
The loss is then back-propagated through all intermediate steps of the classifier adaptation (lines 11 and 12), and is used to update the meta-learner weights $\theta$ (line 14). Therefore, instead of having a feature representation that is directly applicable for any user, they are learned to work well for new users \emph{after} $K$ gradient descent steps on the user's signatures. For stability during training, we train on ``mini-batches" of episodes, by accumulating the gradients for $M$ episodes before updating $\theta$. 

Figure \ref{fig:inner_optimization} illustrates the classifier adaptation procedure. In this work, we adapt the MAML algorithm to use different loss functions for the classifier adaptation and the final loss (used for the meta-update). In particular, we consider a loss function $L$ that only uses genuine signatures for the classifier adaptation, and a loss function $L'$ that use both genuine signatures and forgeries. Let $\mathcal{D}_u = G_u \cup G_{i \neq u}$ be the training set consisted of genuine signatures from the user ($G_u$) and random forgeries ($G_{i \neq u}$). We consider the following loss for classifier adaptation:


\begin{equation}
\begin{aligned}
L(\mathcal{D}_u, \theta) = - \frac{1}{|G_u|} \sum_{x: G_u}{\log(P(y | x, \theta))} \\- \frac{1}{|G_{i \neq u}|} \sum_{x: G_{i \neq u}}{\log(P(y | x, \theta))}
\end{aligned}
\end{equation}
where $|G_u|$ and $|G_{i \neq u}|$ are the number of users in the sets, which is used to correct for the imbalance between the two classes.

Let $\mathcal{D'}_u = G'_u \cup G'_{i \neq u} \cup S'_u$ be the a disjoint set of signatures for user $u$: genuine signatures ($G'_u$), random forgeries ($G'_{i \neq u}$), and (if available), skilled forgeries $S'_u$. We define the loss function for meta-update as follows:

\begin{equation}
\begin{aligned}
L'(\mathcal{D}'_u, \theta) = - \frac{1}{|G'_u|} \sum_{x: G'_u}{\log(P(y | x, \theta_K^{(u)}))} \\ - \frac{1}{|G'_{i \neq u}|} \sum_{x: G'_{i \neq u}}{\log(P(y | x, \theta_K^{(u)}))} \\ - \frac{1}{|S'_u|} \sum_{x: S'_u}{\log(P(y | x, \theta_K^{(u)}))}
\end{aligned}
\end{equation}


\begin{algorithm}
	\caption{Classifier adaptation}
	\label{alg:classifier_adaptation}
	\begin{algorithmic}[1]	
		\Require $K$: Number of gradient descent steps
		\Require $\alpha$ Learning rate
		\Require $\theta$ Meta-learned weights
		\Require $\mathcal{D}_u$ Reference set for user $u$
		\Ensure $\theta^\prime_K$: Weights adapted to the user after K steps
		\State $\theta^\prime_0 \gets \theta$ 
		\For {$k \gets 1$ to $K$} 
		\State $\theta^\prime_k \gets \theta^\prime_{k-1} - \alpha \nabla_{\theta^\prime_{k-1}}{L(\mathcal{D}_u, \theta^\prime_{k-1})}$ 
		\EndFor	
	\end{algorithmic}
\end{algorithm}

On generalization, for a new user we first adapt the weights to this user using a set of reference signatures $\mathcal{D}_u$, and then classify a new query signature using the adapted weights. Algorithm \ref{alg:classifier_adaptation} describes the classifier adaptation to a new user. We note that only the loss function $L$ is used, and therefore only genuine signatures are used when adapting a classifier for a new user.

\subsection{Meta-learning for one-class classification}

The approach defined above can also be extended for one-class classification, where the classifier adaptation is done with only genuine signatures from the user of interest. This is easily implemented by considering $\mathcal{D}_u = G_u$. 
It is worth noting that similarity-based methods and one-class methods that involve feature learning often suffer from the problem of collapsing representations into a point \cite{perera_learning_2018}. This is often addressed by adding a penalty in the loss function that requires dissimilar items to be far apart in the feature space. In our formulation, while the user's classifier is only trained with data from one class, we observe that training does not collapse to a single point since the meta-training procedure directly optimizes the performance on separating forgeries in $\mathcal{D}^\prime_u$.

\section{Experimental Protocol}
\label{sec:experimental_protocol}

We conducted most experiments on the GPDS-960 dataset \cite{vargas_off-line_2007}, that consists of 881 users, with 24 genuine signatures per user and 30 skilled forgeries. We follow the same dataset separation as previous work (figure \ref{fig:dataset_separation}), with users 350-881 as $\mathcal{D}_\text{meta-train}$, 300-350 as $\mathcal{D}_\text{meta-val}$ and users 0-300 as $\mathcal{D}_\text{meta-test}$. We used the same pre-processing method from previous work \cite{hafemann_ijcnn_2016, hafemann_pr_2017}, by removing the background noise using OTSU, centering the images in a canvas of size $952 \times 1360$ and resizing them to $170 \times 242$.

We analyze the impact of the hyperparameters in the classifier's performance, measured in $\mathcal{D}_\text{meta-val}$. We consider the experiments by varying these parameters:

\begin{itemize}
	\item Number of gradient descent steps in the classifier adaptation: $K \in \{1, 5\}$
	\item One-class classification vs adaptation using genuine signatures and random forgeries
	\item Fraction of users with skilled forgeries available for training
	\item Performance as we vary the number of reference genuine signatures 
\end{itemize}

We compare the results on $\mathcal{D}_\text{meta-val}$ with a baseline using feature learning followed WD classification \cite{hafemann_pr_2017}. As in \cite{hafemann_pr_2017}, we evaluate each model with repeated random subsampling: we randomly partition the validation set into training ($\mathcal{D}_u$) and testing ($\mathcal{D}'_u$), repeating the experiment 10 times with different partitions. We report the mean and standard deviation of the metrics.

In all experiments, we train the meta-classifier for a total of 100 epochs, considering a meta-batch size $M=4$. We consider an initial meta-learning rate $\beta = 0.001$, that is reduced (with cosine annealing) to $10^{-5}$ by the last epoch. We used early stopping, by keeping the meta-learner weights that performed best in the validation set. Following \cite{antoniou_how_2018}, we used Multi-Step Loss Optimization (MSL) for improving training stability. For the first 20 epochs, instead of computing the loss function $L'$ only after $K$ steps (step 12 of algorithm \ref{alg:metalearning}), we compute the loss function for all intermediate $\theta'_k$, and consider a weighted average of the losses. In the first epoch the loss using each $\theta'_k$ contributes equally to the loss function, and the weights are annealed to give more weight to the last step until iteration 20, after which only the loss function at the final step $K$ contributes to the loss. We found this procedure effective in stabilizing training (measured by the variation in validation accuracy across epochs). We also attempted to use learnable task learning rates (LSLR) described in \cite{antoniou_how_2018} without success. Empirically, we also noticed that when using only genuine signatures the task learning rate needs to be larger than the case where skilled forgeries are available for training. In our experiments, if the fraction of users with skilled forgeries is less than 10\% we used a task learning rate $\alpha = 0.01$, and a learning rate of $\alpha = 0.001$ for the other experiments.

In order to evaluate the transferability of the features to other operating conditions, we conducted experiments on other datasets, (that were collected in different regions, and followed different collection processes): MCYT-75 \cite{ortega-garcia_mcyt_2003}, CEDAR \cite{kalera_offline_2004} and Brazilian PUC-PR \cite{freitas_bases_2000}. We conducted two experiments: (i) use the meta-learner trained on GPDS directly for new users of these datasets; (ii) train a meta-learner with data from the four datasets. It is worth noting that, with the exception of GPDS, the datasets are relatively small, with 55, 75 and 60 users for CEDAR, MCYT and Brazilian PUC-PR. We observed that the formulations from this work require a large amount of users for training, and for this reason, we conducted 10-fold cross validation. We divide each dataset in 10 folds (by users), and for each run we consider 1 fold as meta-test, and the remaining folders for meta-training and validation. As in the previous experiments, we further use repeated subsampling for evaluating the adaptation for the new users. In total, for experiment (ii), we train 10 CNN models and perform 10 adaptations for each user. We report the mean error rates over all runs, and the standard deviation across the 10 different adaptations (each based on different train/test splits of the repeated subsampling).

\begin{table} \centering
	\caption{Base architecture used in this work}
	\begin{tabular}{@{}lc@{}} \hline
		Layer & Size \\ \hline
		
		Input & 1x150x220 \\
		Convolution (C1)& 32x5x5 \\
		Max Pooling & 32x5x5 \\
		
		Convolution (C2)& 32x5x5 \\
		Pooling & 32x5x5 \\
		
		Fully Connected (FC3)& 1024 \\
		Fully Connected (FC4)& 256 \\
		Fully Connected + Sigmoid  & 1  \\

		\hline
	\end{tabular}
	\label{table:cnn_architecture}
\end{table}

The CNN architecture used in the experiments is listed in table \ref{table:cnn_architecture}. We found that using a smaller network, compared to previous work using feature learning followed by WD classification, was successful in the meta-learning setting. This network has a total of 1.4M weights and uses 0.1 GFLOPS for forward propagation, while SigNet \cite{hafemann_pr_2017} has 15.8M weights and uses 0.6 GFLOPS. That is, the CNN used in this work is 10x smaller and 6x times faster. 

We evaluate the performance using the following metrics: False Rejection Rate (FRR): the fraction of genuine signatures rejected as forgeries; False Acceptance Rate (FAR\textsubscript{random} and FAR\textsubscript{skilled}): the fraction of forgeries accepted as genuine (considering random forgeries and skilled forgeries). We also report the Equal Error Rate (EER): which is the error when FAR = FRR. We considered two forms of calculating the EER: EER\textsubscript{global $\tau$}: using a global decision threshold and EER\textsubscript{user $\tau$}: using user-specific decision thresholds. In both cases, to calculate the Equal Error Rate we only considered skilled forgeries. For FRR and FAR, we report the values with a threshold of $0.5$ (i.e. if $p(y=1 | x, \theta'_K) \ge 0.5$ we consider the model predicting $x$ as a genuine signature).

\begin{table*}
	\centering
	\caption{Performance on $\mathcal{D}_\text{meta-val}$ with one-class and two-class formulations}
	\label{tbl:one_vs_twoclass}
	\begin{tabular}{llllllll}
		\toprule
		Type & \#Gen &\#RF &               FRR &        FAR\textsubscript{random} &       FAR\textsubscript{skilled} &    EER\textsubscript{global $\tau$} & EER\textsubscript{user $\tau$} \\
		\midrule
		SigNet* + WD   & 5 & 7434 &  10.48 ($\pm 2.24$) &  0.03 ($\pm 0.01$) &  24.67 ($\pm 0.99$) &  17.03 ($\pm 1.06$) &  13.17 ($\pm 0.94$) \\
		SigNet-F* + WD & 5 &  7434 &  18.08 ($\pm 1.49$) &  0.16 ($\pm 0.04$) &   1.55 ($\pm 0.22$) &    4.6 ($\pm 0.59$) &   3.08 ($\pm 0.38$) \\ \midrule
		
		\makecell[l]{Meta-learning \\ One-class} & 5 &-  &  2.54 ($\pm 0.61$) &  2.74 ($\pm 0.93$) &  4.24 ($\pm 0.93$) &  3.48 ($\pm 0.57$) &   1.69 ($\pm 0.43$) \\ \cmidrule{2-8}
		\multirow{4}{*}{\makecell[l]{Meta-learning \\ Two-class}} & 5  & 5  &  2.82 ($\pm 0.59$) &  1.98 ($\pm 0.55$) &  4.18 ($\pm 0.48$) &   3.8 ($\pm 0.48$) &   2.04 ($\pm 0.41$) \\
		& 5 & 10 &   5.1 ($\pm 0.99$) &  1.94 ($\pm 0.27$) &  2.66 ($\pm 0.76$) &  3.56 ($\pm 0.57$) &   1.85 ($\pm 0.57$) \\
		& 5 & 20 &  2.84 ($\pm 0.97$) &  1.98 ($\pm 0.48$) &   3.1 ($\pm 0.83$) &  2.86 ($\pm 0.59$) &   1.78 ($\pm 0.27$) \\
		& 5 & 30 &  2.62 ($\pm 0.82$) &  2.48 ($\pm 0.39$) &  3.46 ($\pm 0.62$) &  3.17 ($\pm 0.37$) &    1.4 ($\pm 0.48$) \\
		\bottomrule
	\end{tabular}
\end{table*}

\section{Results}
\label{sec:results}

\subsection{System design}
\label{sec:results_systemdesign}

In this section we report the results on $\mathcal{D}_\text{meta-val}$ (GPDS users 300-350), considering the experiments defined in section \ref{sec:experimental_protocol}. The objective is to evaluate different aspects of the system, such as the number of gradient steps (that trades-off computation complexity and accuracy), as well as investigate the performance of the model in different data scenarios.

\begin{figure}
	\centering
	\includegraphics[width=0.8\columnwidth]{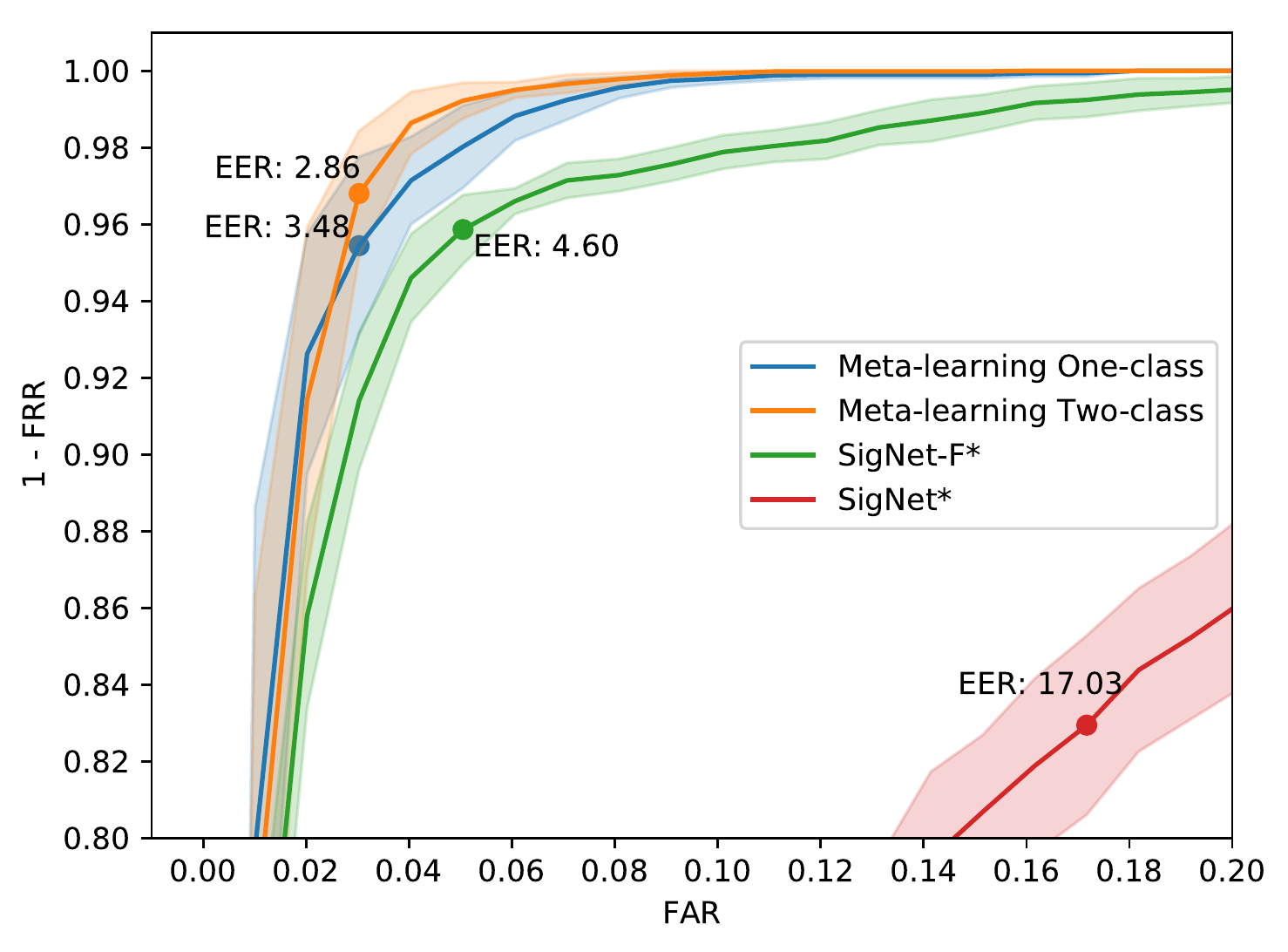}
	\caption{ROC curves on $\mathcal{D}_\text{meta-val}$ comparing the one-class and two-class formulations with the baselines. }
		\label{fig:roc}
\end{figure}

In a first experiment we consider the results of the one-class formulation and the two-class formulation as we vary the number of Random Forgeries used for classifier adaptation (\#RF). For this experiment, use used 5 genuine signatures for classifier adaptation, and $K=5$ gradient descent steps; for meta-training we considered that skilled forgeries were available on $\mathcal{D}_\text{meta-train}$ (users 350-881). Note that for validation, no skilled forgeries were used for training. Table \ref{tbl:one_vs_twoclass} reports the results of these experiments. We observe similar verification performance on the two formulations. Note that the formulation using random forgeries is more computationally expensive, as the classifier adaptation involves a larger batch of images (e.g. computing the loss for one-class uses 5 images, while for two-class with \#RF=30 uses 35 images). We also compare with a method using feature learning followed by WD classification \cite{hafemann_pr_2017}. The entries denoted \emph{SigNet*} used the same approach proposed in \cite{hafemann_pr_2017}, but using the CNN architecture defined for this work (table \ref{table:cnn_architecture}). We note that the meta-learning formulation performed much better, while being a simpler model (single model for all users). A comparison with the SigNet CNN architecture from \cite{hafemann_pr_2017} is conducted in section \ref{sec:soa}, where we compare to the state-of-the-art. Figure \ref{fig:roc} presents ROC curves for the one-class formulation and the two-class formulation with \#RF=20, as well as the two baselines. We consider average ROC curves over the 10 folds, where the solid line indicates the mean FRR for a given value of FAR, and the shaded area indicates one standard-deviation. Again, we see improved performance compared to the baselines. 

\begin{figure}
	\centering
	\includegraphics[width=0.8\columnwidth]{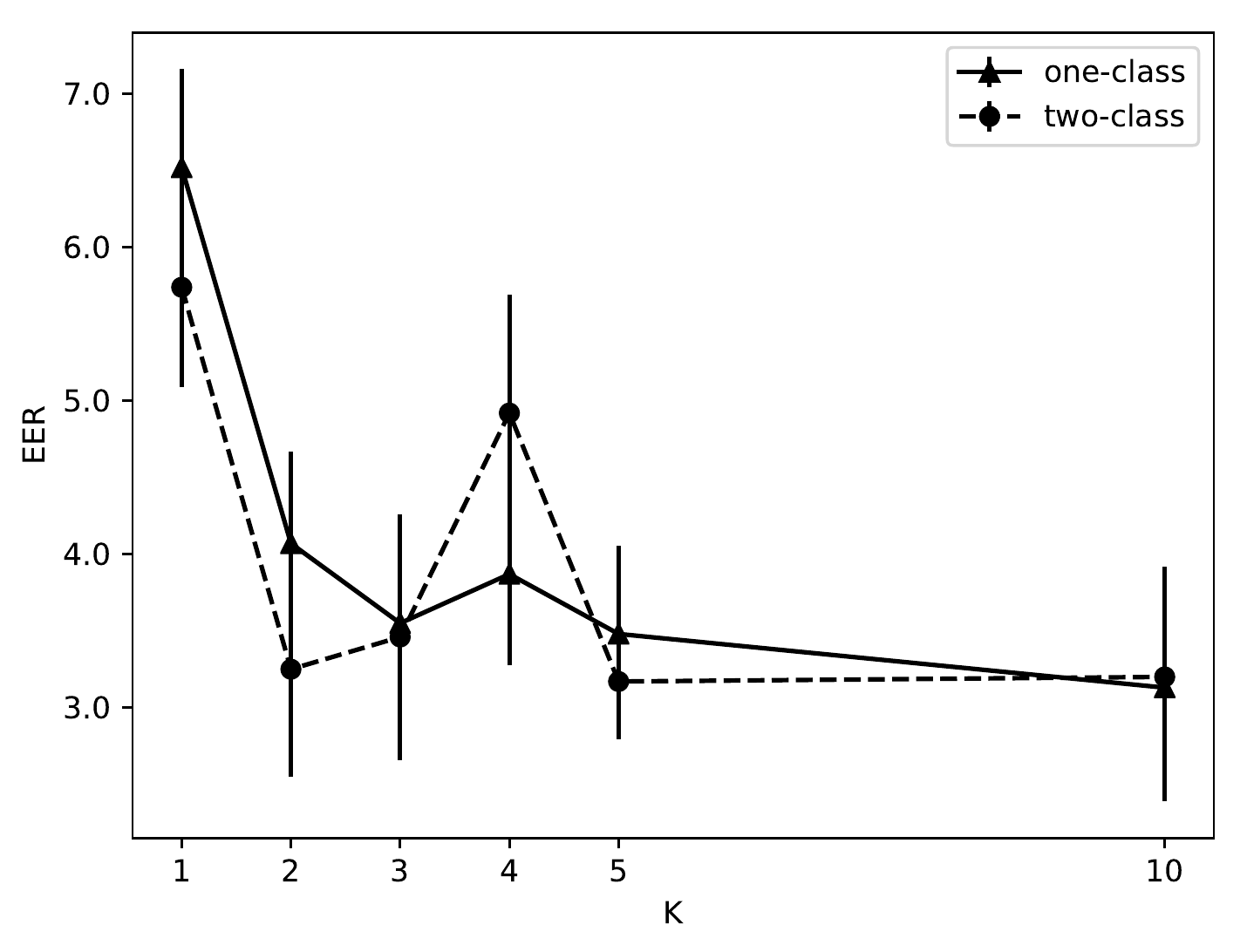}
	\caption{Performance on $\mathcal{D}_\text{meta-val}$ as we vary the number of update steps $K$.}
	\label{fig:varying_k}
\end{figure}

Figure \ref{fig:varying_k} shows the results on verification performance as we vary the number of gradient descent steps $K$. For each value of $K$, we meta-trained a network and evaluate its performance on $\mathcal{D}_\text{meta-val}$. We observed improved performance with larger number of steps, but with diminishing returns. We note a high variance of the errors in these experiments, and therefore we cannot determine a particular $K$ as being optimal. As we increase the number of steps, we also increase the computational cost. If we consider that forward propagation and backward propagation have similar cost, the classifier adaptation for a new user takes $2K$ the time for a single forward pass. A higher $K$ also requires more memory (in the order of $2K$) during meta-training, since the whole update sequence needs to be stored in memory in order to compute the gradient for meta-update (as can be seen in figure \ref{fig:inner_optimization}).

\begin{figure}
	\centering
	\includegraphics[width=0.8\columnwidth]{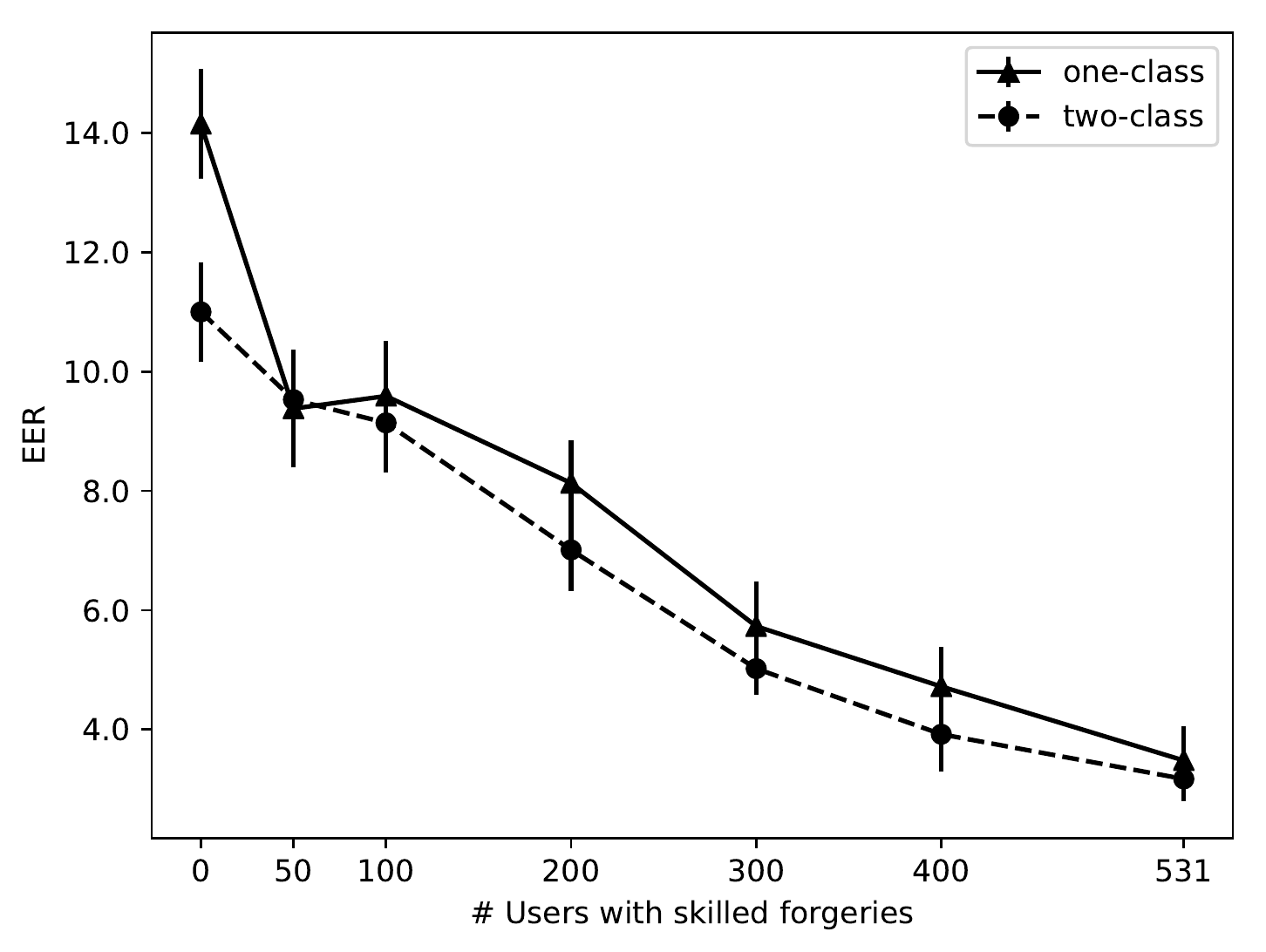}
	\caption{Performance on $\mathcal{D}_\text{meta-val}$ as we vary the number of users in $\mathcal{D}_\text{meta-train}$ for which skilled forgeries are available.}
	\label{fig:varying_sksize}
\end{figure}

\begin{figure}
	\centering
	\subfloat[One-class]{	\includegraphics[width=0.8\columnwidth]{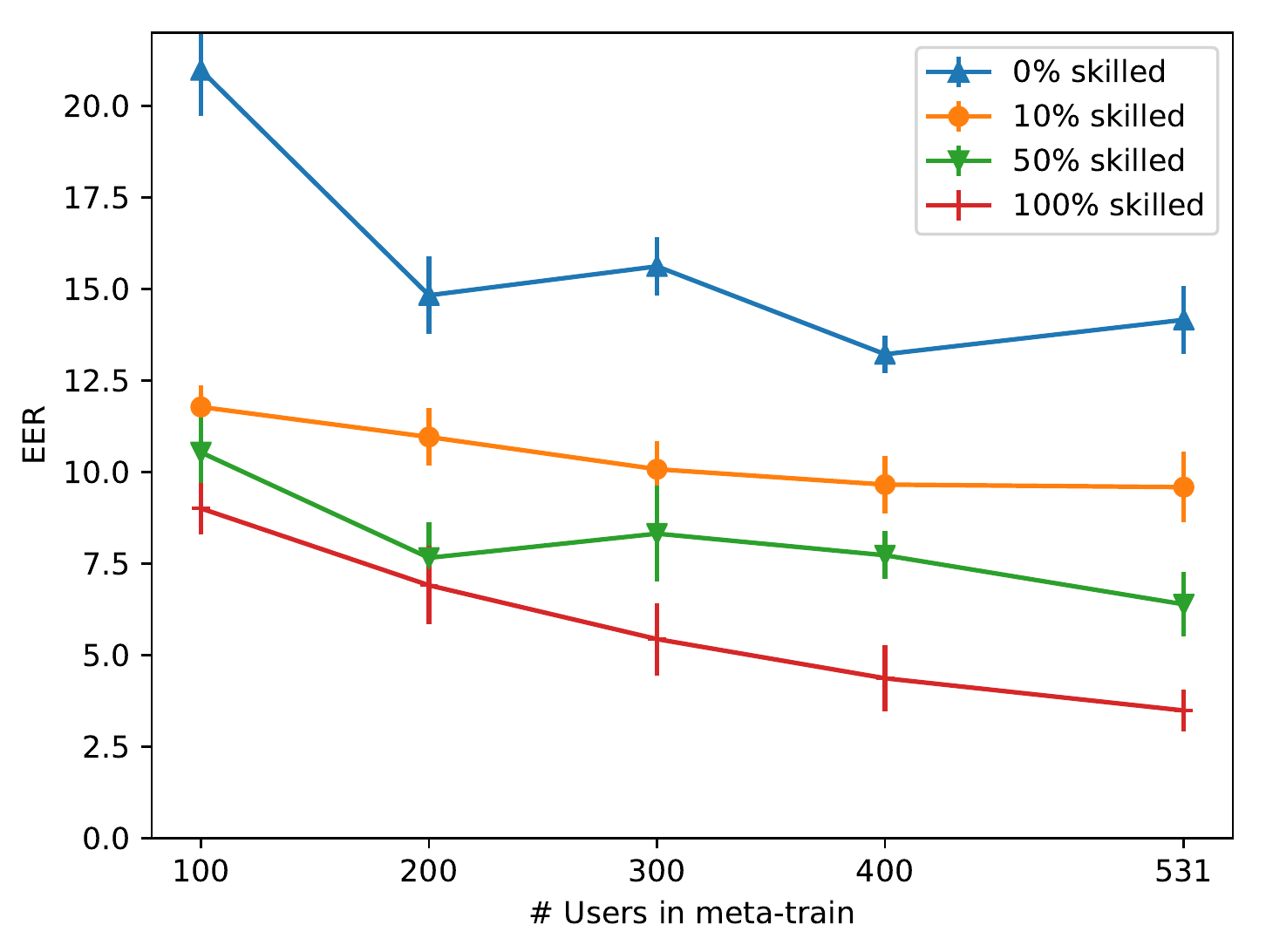}}
	\qquad
	\subfloat[Two-class]{	\includegraphics[width=0.8\columnwidth]{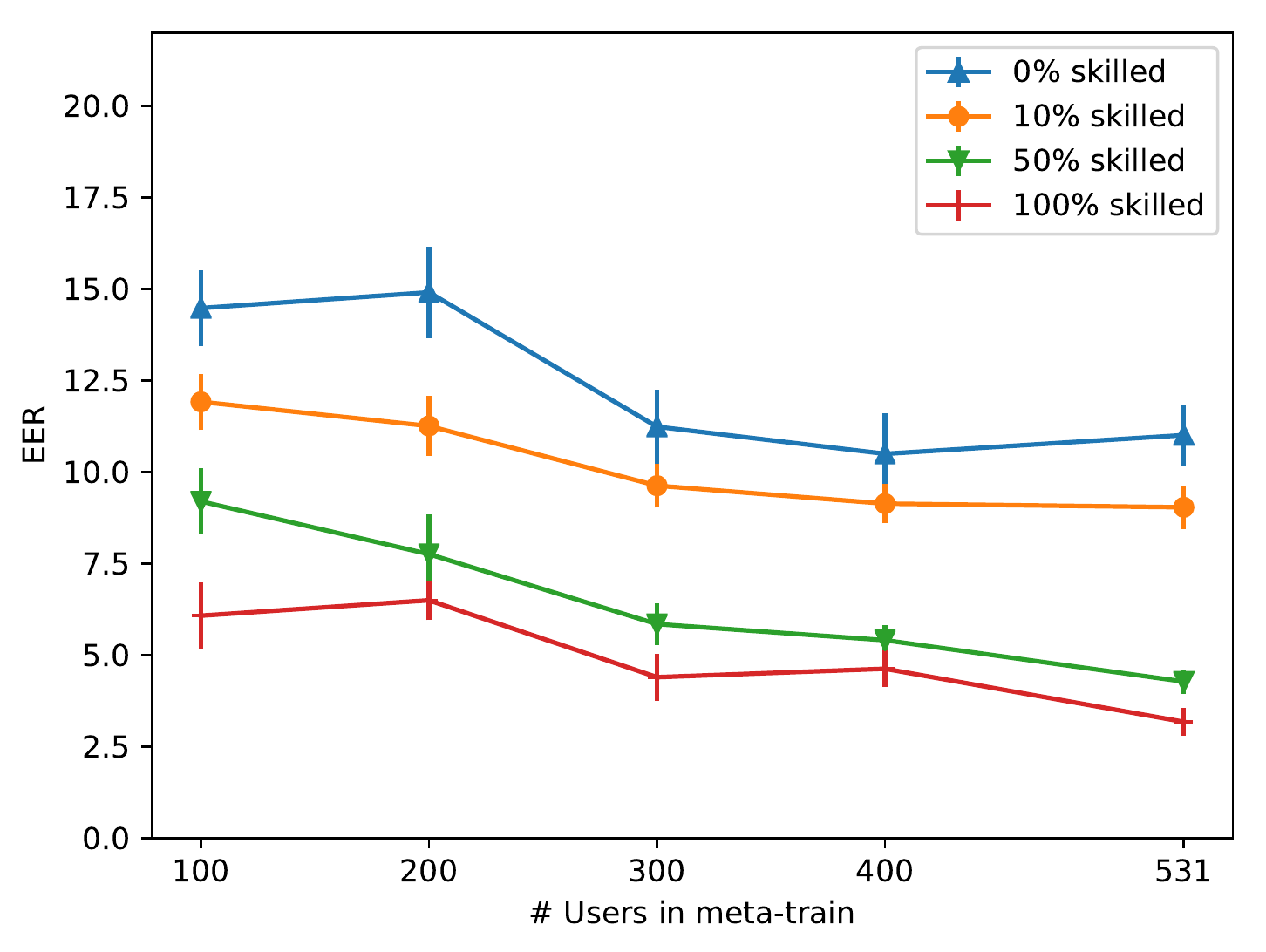}
	\label{fig:varying_devsize_twoclass}}
	\caption{Performance on $\mathcal{D}_\text{meta-val}$ as we vary the number of users available for meta-training. (a): one-class formulation; (b) two-class formulation.}
	\label{fig:varying_devsize}
\end{figure}

In figures \ref{fig:varying_sksize} and \ref{fig:varying_devsize} we analyze the impact in performance as we vary the size of the $\mathcal{D}_\text{meta-train}$ set. 
As noted in section \ref{sec:maml_for_sigver}, if skilled forgeries from a subset of users are available, we can incorporate them into the meta-update loss function $L'$.
In this experiment we considered that $\mathcal{D}_\text{meta-train}$ contains all 531 users, and vary the number of users for which skilled forgeries are available. For each case, we build a dataset consisting of genuine signatures for all users and skilled forgeries for the selected users, and trained a model. Figure \ref{fig:varying_sksize} shows the performance as we vary the number of users for which skilled forgeries as available. We re-iterate that we evaluate the performance on a disjoint set of users ($\mathcal{D}_\text{meta-val}$) for which only genuine signatures are used. We observed that the meta-learning formulation of the problem is well suited to incorporating information from skilled forgeries (when it is available), and this generalizes well to unseen users, for which we only have genuine signatures. However, we observed that the performance is not very good when there are only genuine signatures for meta-training: the one-class formulation achieves 14.15\% EER when only genuine signatures are available, and 3.48\% EER when skilled forgeries are available for all 531 users in meta-training.

In figure \ref{fig:varying_devsize}, we evaluate the performance of the system as we vary the number of users in $\mathcal{D}_\text{meta-train}$. We also consider 4 levels of availability of skilled forgeries in the meta-training set: 0\% (genuine only), 10\%, 50\% and 100\%, where the percentages refer to the number of users for which skilled forgeries are available (e.g. 10\% with 100 users means that forgeries for 10 users are considered, where the remaining 90 users have only genuine signatures). For a given number of users and skilled forgery percentage, we construct a dataset with randomly selected users (taken from the 531 users in the development set), with genuine signatures from all the selected users, and skilled forgeries for a fraction of the users. We then use this dataset for meta-training a model, and evaluate its performance on $\mathcal{D}_\text{meta-val}$. We observed improved performance both as more users are available for meta-training, as well as when more knowledge of skilled forgeries is available. Most surprisingly, we observed that for the two-class formulation, a classifier trained with 100 users with 100\% forgeries (i.e. forgeries for every user in meta-train) performed better than a model trained with 531 users with forgeries for only 100 users (comparing figures \ref{fig:varying_devsize_twoclass} and \ref{fig:varying_sksize}): 6.07\% EER vs 9.14\% EER. We re-iterate that this measures the performance on discriminate genuine signatures and skilled forgeries, and the model that has access to more users (with the same amount of users with skilled forgeries) has better performance on discriminating random forgeries, since its optimization consisted mostly of this problem.

\subsection{Comparison with the state-of-the-art}
\label{sec:soa}

\begin{table}
	\centering
	\caption{Comparison with state-of-the art on the GPDS dataset (errors in \%)}
	\label{tbl:soa_gpds}
	\resizebox{\columnwidth}{!}{%
		\begin{threeparttable}
			
			\begin{tabular}{lllllr}
				\hline
				Reference & Type &Dataset& \begin{tabular}[x]{@{}c@{}}\#samples\\per user\end{tabular}  & Features &  EER\\
				\hline
				Hu and Chen \cite{hu_offline_2013}& WI & GPDS-150 &10 &LBP, GLCM, HOG & 7.66\\
				Guerbai et al \cite{guerbai_effective_2015} &WD & GPDS-160&12&Curvelet transform&15.07\\
				Serdouk et al \cite{serdouk_new_2015} & WD &GPDS-100 & 16 & GLBP, LRF& 12.52 \\
				Yilmaz \cite{yilmaz_score_2016}  &WD &GPDS-160 &5&LBP, HOG, SIFT& 7.98\\
				Yilmaz \cite{yilmaz_score_2016}  &WD &GPDS-160 &12&LBP, HOG, SIFT& 6.97\\
				Soleimani et al \cite{soleimani_deep_2016}  &WI &GPDS-300 &10&LBP& 20.94\\
								
				
				Hafemann et al \cite{hafemann_pr_2017}  & WD &GPDS-300 & 5  & SigNet-F (global $\tau$)  & 5.25 ($\pm 0.15$)\\
				Hafemann et al \cite{hafemann_pr_2017}  & WD &GPDS-300 & 5  & SigNet-F (user $\tau$) & 2.42 ($\pm 0.24$)\\
				
				Hafemann et al \cite{hafemann_pr_2017}  & WD &GPDS-300 & 12  & SigNet-F (global $\tau$) & 3.74 ($\pm 0.15$)\\
				Hafemann et al \cite{hafemann_pr_2017}  & WD &GPDS-300 & 12  & SigNet-F (user $\tau$)& 1.69 ($\pm 0.18$)\\
				
				Souza et al \cite{souza_writer-independent_2018}  & WI &GPDS-300 & 5 & SigNet (global $\tau$)& 9.05 ($\pm 0.34$) \\
				Souza et al \cite{souza_writer-independent_2018}  & WI &GPDS-300 & 5 & SigNet (user $\tau$)& 4.40 ($\pm 0.34$) \\
				Souza et al \cite{souza_writer-independent_2018} &WI & GPDS-300 & 12 & SigNet (global $\tau$)&7.96 ($\pm 0.26$)\\
				Souza et al \cite{souza_writer-independent_2018} &WI & GPDS-300 & 12 & SigNet (user $\tau$)&3.34 ($\pm 0.22$)\\ \midrule
				
				Present work & WI/WD &  GPDS-300 & 5 & MAML one-class (global $\tau$) & 5.52 ($\pm 0.20$)\\
				
				Present work & WI/WD &  GPDS-300 & 5 & MAML one-class (user $\tau$) & 3.35 ($\pm 0.13$)\\
				
				Present work & WI/WD &  GPDS-300 & 5 & MAML two-class (global $\tau$) & 5.16 ($\pm 0.19$)\\
				
				Present work & WI/WD &  GPDS-300 & 5 & MAML two-class (user $\tau$) & 2.94 ($\pm 0.20$)\\
				
				Present work & WI/WD &  GPDS-300 & 12 & MAML one-class (global $\tau$) & 4.70 ($\pm 0.11$)\\
				Present work & WI/WD &  GPDS-300 & 12 & MAML one-class (user $\tau$) & 2.93 ($\pm 0.27$)\\
				
				Present work & WI/WD &  GPDS-300 & 12 & MAML two-class (global $\tau$) & 4.39 ($\pm 0.18$)\\
								
				Present work & WI/WD &  GPDS-300 & 12 & MAML two-class (user $\tau$) & 2.68 ($\pm 0.17$)\\
				
				\hline
			\end{tabular}
		\end{threeparttable}
	}
\end{table}

We now compare our results with the state-of-the-art in the GPDS-300 dataset. For these comparisons, we considered a model trained with the one-class formulation, and a model trained with the two-class formulation, with $r=30$ forgeries. In both cases, we used the whole dataset $\mathcal{D}_\text{meta-train}$ for training the meta-classifier, and used 5 genuine signatures for classifier adaptation, with $k=5$ updates. While training was conducted with 5 reference signatures, we evaluate the performance of the system with different number of references.

Table \ref{tbl:soa_gpds} compares our results with the state-of-the-art. We observe an improved performance compared to other WI systems, achieving 5.16\% EER (global $\tau$) with 5 reference signatures, compared to 9.05\% from \cite{souza_writer-independent_2018}. Comparing to WD systems, we observed similar performance in some scenarios (5 reference signatures), and worse results otherwise. With 12 reference signatures, the proposed system obtained 4.39\% EER (global $\tau$), compared to 3.74 for the WD system \cite{hafemann_pr_2017}. However, the proposed system is more scalable, as a single model is stored for all users.

\begin{figure}
	\centering
	\subfloat[One-class]{	\includegraphics[width=0.8\columnwidth]{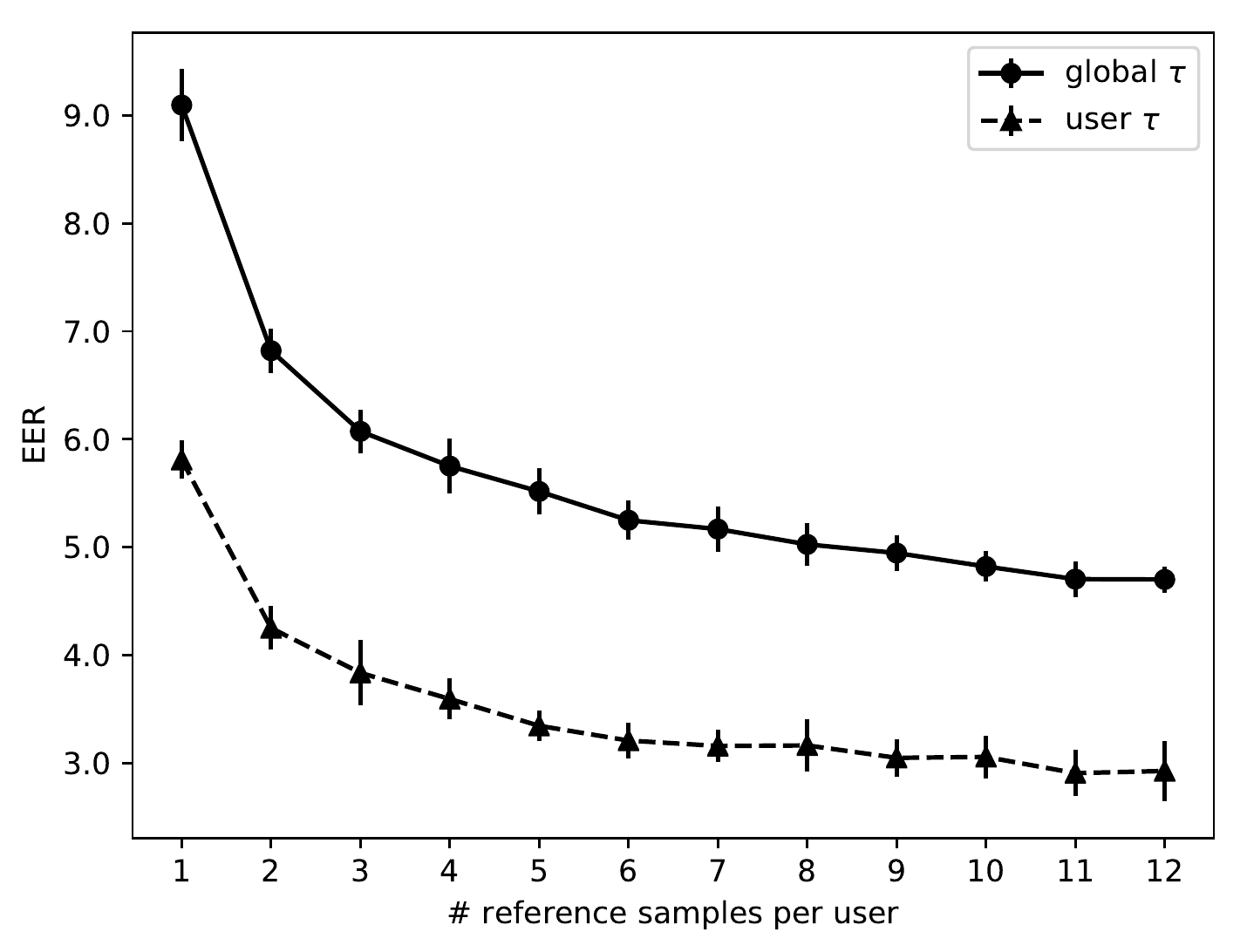}}
	\qquad
	\subfloat[Two-class]{	\includegraphics[width=0.8\columnwidth]{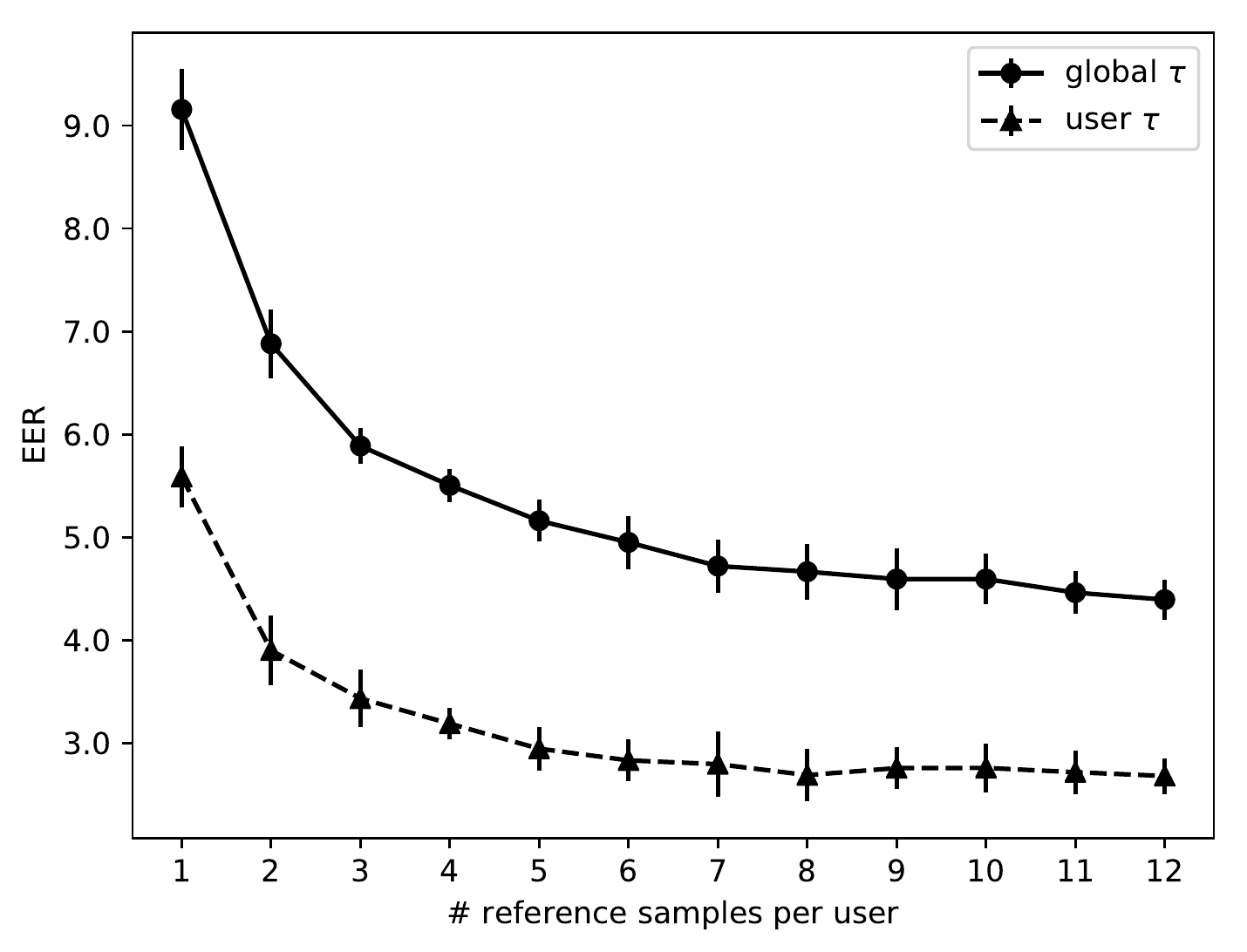}}
	\caption{Performance on GPDS-300 as we vary the number reference signatures available for each user. (a): one-class formulation; (b) two-class formulation.}
	\label{fig:varying_nref}
\end{figure}

Figure \ref{fig:varying_nref} shows the performance on GPDS-300 as we vary the number of reference samples available for each user. As commonly observed in WD systems (e.g. \cite{hafemann_pr_2017}), the performance greatly improves as more reference samples are available for training: For the one-class formulation, performance with a single reference is 9.09\% EER (global $\tau$) and 5.81\% EER (user $\tau$). With 12 references, we obtain 4.70\% EER (global $\tau$) and 2.93\% EER (user $\tau$).

\subsection{Transfer to other datasets}

\begin{table}
	\centering
	\caption{Transfer performance to the other datasets}
	\label{tbl:transfer}
	\begin{tabular}{lllll}
		\toprule
		Target Dataset & Training Dataset & EER (global)    & EER (user)      &  \\ \midrule
		MCYT           & GPDS     & 15.48 ($\pm$ 1.00) & 12.54 ($\pm$ 1.86) &  \\
		& All      & 15.37 ($\pm$ 0.97) & 12.77 ($\pm$ 0.46) &  \\ \midrule
		CEDAR          & GPDS     & 15.98 ($\pm$ 1.09) & 12.07 ($\pm$ 1.01) &  \\
		& All      & 10.69 ($\pm$ 1.76) & 8.02 ($\pm$ 1.22)  &  \\ \midrule
		Brazilian      & GPDS     & 8.05 ($\pm$ 0.95)  & 4.83 ($\pm$ 1.07)  &  \\
		& All      & 8.55 ($\pm$0.55)   & 6.7 ($\pm$ 0.87)   & \\ \bottomrule
	\end{tabular}	
\end{table}

\begin{table}
	\centering
	\caption{Comparison with the state-of-the-art in MCYT}
	\label{tbl:mcyt_soa}
	\resizebox{\columnwidth}{!}{%
		\begin{tabular}{llrlc}
			\hline
			Reference & Type & \# Samples & Features & EER\\
			\hline
			Wen et al.\cite{wen_model-based_2009}&WD&5&RPF &15.02\\
			Vargas et al.\cite{vargas_off-line_2011}&WD&5&LBP &11.9\\
			Vargas et al.\cite{vargas_off-line_2011}&WD&10&LBP &7.08\\
			Ooi et al\cite{ooi_image-based_2016}&WD&5&DRT + PCA&13.86\\
			Ooi et al\cite{ooi_image-based_2016}&WD&10&DRT + PCA&9.87\\
			Soleimani et al.\cite{soleimani_deep_2016}&WD&5&HOG &13.44\\
			Soleimani et al.\cite{soleimani_deep_2016}&WD&10&HOG&9.86\\
			
			Hafemann et al.\cite{hafemann_pr_2017} & WD & 5 & SigNet (user $\tau$) & 3.58 ($\pm$ 0.54) \\
			Hafemann et al.\cite{hafemann_pr_2017} & WD & 10 & SigNet (user $\tau$) & 2.87 ($\pm$ 0.42) \\
			\hline
			Present Work & WI/WD & 5 & MAML one-class (global $\tau$) & 15.37($\pm$ 0.97) \\
			Present Work & WI/WD & 5 & MAML one-class (user $\tau$) & 12.77($\pm$ 0.46) \\
			
			Present Work & WI/WD & 10 & MAML one-class (global $\tau$) & 14.50($\pm$ 0.77) \\
			Present Work & WI/WD & 10 & MAML one-class (user $\tau$) & 12.44($\pm$ 0.97) \\

			\hline
		\end{tabular}
	}
\end{table}

\begin{table}
	\centering
	\caption{Comparison with the state-of-the-art in CEDAR}
	\label{tbl:cedar_soa}
	\resizebox{\columnwidth}{!}{%
		
		\begin{tabular}{llrlc}
			\hline
			Reference & Type &\# Samples & Features & AER/EER\\
			\hline
			Chen and Srihari\cite{chen_new_2006}&WD&16&Graph Matching&7.9\\
			Kumar et al.\cite{kumar_writer-independent_2010}&WI&1&morphology &11.81\\
			Kumar et al.\cite{kumar_writer-independent_2012}&WI&1&Surroundness &8.33\\
			Bharathi and Shekar\cite{bharathi_off-line_2013}&WD&12&Chain code &7.84\\
			Guerbai et al.\cite{guerbai_effective_2015}&WD & 4&Curvelet transform&8.7\\
			Guerbai et al.\cite{guerbai_effective_2015}&WD & 8&Curvelet transform&7.83\\
			Guerbai et al.\cite{guerbai_effective_2015}&WD & 12&Curvelet transform&5.6\\
			
			Hafemann et al.\cite{hafemann_pr_2017} & WD & 4 & SigNet (SVM) & 5.87 ($\pm$ 0.73) \\
			Hafemann et al.\cite{hafemann_pr_2017} & WD & 8 & SigNet (SVM) & 5.03 ($\pm$ 0.75) \\
			
			\hline
			Present Work & WI/WD & 4 & MAML one-class (global $\tau$) & 11.06($\pm$ 1.12) \\
			Present Work & WI/WD & 4 & MAML one-class (user $\tau$) & 8.27($\pm$ 1.45) \\
			
			Present Work & WI/WD & 8 & MAML one-class (global $\tau$) & 10.21($\pm$ 1.21) \\
			Present Work & WI/WD & 8 & MAML one-class (user $\tau$) & 7.07($\pm$ 1.08) \\
			\hline
		\end{tabular}
	}
\end{table}

\begin{table}
	\centering
	\caption{Comparison with the state-of-the-art on the Brazilian PUC-PR dataset (errors in \%)}
	\label{tbl:brazilian_soa}
	\resizebox{\columnwidth}{!}{%
		
		\begin{tabular}{llllr}
			\hline
			Reference & Type & \#samples & Features & AER\textsubscript{genuine + skilled}/EER   \\
			\hline
			Bertolini et al. \cite{bertolini_reducing_2010} &WI& 15 &Graphometric & 8.32  \\
			Batista et al. \cite{batista_dynamic_2012} & WD&30 & Pixel density &10.5\\
			Rivard et al. \cite{rivard_multi-feature_2013} & WI & 15 &ESC + DPDF &11.08\\
			Eskander et al. \cite{eskander_hybrid_2013}& WD& 30 &ESC + DPDF &10.67\\

			Hafemann et al.\cite{hafemann_pr_2017} & WD&                5 &               SigNet (user $\tau$) &  2.92 ($\pm$ 0.44) \\
			Hafemann et al.\cite{hafemann_pr_2017} & WD&               15 &               SigNet (user $\tau$) &  2.07 ($\pm$ 0.63) \\
			
			Souza et al.\cite{souza_writer-independent_2018} & WI&                5 &               SigNet (global $\tau$) &  5.95 ($\pm$ 0.68) \\
			Souza et al.\cite{souza_writer-independent_2018} & WI&                5 &               SigNet (user $\tau$) &  2.58 ($\pm$ 0.72) \\
			Souza et al.\cite{souza_writer-independent_2018} & WI&                15 &               SigNet (global $\tau$) &  5.13 ($\pm$ 0.23) \\
			Souza et al.\cite{souza_writer-independent_2018} & WI&                15 &               SigNet (user $\tau$) &  1.70 ($\pm$ 0.40) \\			
			
			\hline
			
			Present Work & WI/WD & 5 & MAML one-class (global $\tau$) & 8.55 ($\pm$ 0.55) \\
			Present Work & WI/WD & 5 & MAML one-class (user $\tau$) & 6.70($\pm$ 0.87) \\
			
			Present Work & WI/WD & 15 & MAML one-class (global $\tau$) & 6.93($\pm$ 0.73) \\
			Present Work & WI/WD & 15 & MAML one-class (user $\tau$) & 5.74($\pm$ 0.84) \\
			
			\hline
		\end{tabular}
	}
\end{table}

We now consider results on three other datasets: MCYT, CEDAR and the Brazilian PUC-PR. Table \ref{tbl:transfer} shows the performance in two scenarios: (i) meta-learner trained only in GPDS, with its generalization to new operating conditions and (ii) meta-learned trained on all four datasets (using 10-fold cross validation, as described in section \ref{sec:experimental_protocol}). While the method generalized well to unseen GPDS users, we see that the generalization performance to other datasets is much worse. Furthermore, we notice that even when training with a subset of users from all datasets, the performance does not improve for all datasets. A possible explanation is that the GPDS dataset is still much larger (10 times larger than the others) and dominates training. Overall, this suggests that the proposed method requires a large amount of data from the target application, and is sensitive to changes in operating conditions. Finally, tables \ref{tbl:mcyt_soa}, \ref{tbl:cedar_soa} and \ref{tbl:brazilian_soa} compares de results with the state-of-the-art on MCYT, CEDAR and Brazilian PUC-PR, respectively. 

It is worth noting that the meta-learning does generalize to \emph{new users} of the GPDS dataset, as verified in sections \ref{sec:results_systemdesign} and \ref{sec:soa}, since we evalute in a $\mathcal{D}_\text{meta-test}$ that contains a disjoint set of users that was used to train the meta-learner. What we observed, however, is that this meta-learned does not transfer well to other datasets. This has been observed more recent work with meta-learning \cite{triantafillou_meta-dataset:_2019}, that shows that although these models perform well for new classes of the same distribution (e.g. same dataset), the performance deteriorates when evaluating on new datasets (i.e. a shift in the task-distribution). This is still an active area of research in meta-learning.

\section{Conclusion}
\label{sec:conclusion}

In this paper we proposed to formulate Signature Verification as a meta-learning problem, where each user defines a task. This formulation enables directly optimizing for the objective (separating genuine signatures and forgeries) even when forgeries are not available for all users. The resulting system is scalable and yet adaptable for individual users: a single meta-classifier is learned and stored, and for the verification of a given signature, the classifier is adapted to the claimed user and subsequently used for verification. The proposed method is also able to naturally incorporate new reference signatures for a user, and enable adapting the representation as more training data is available. The drawbacks of this solution are twofold: increased computational cost and worse transferability to new conditions. The method is $2K$ slower, when using $K$ updates for the classification adaptation, although it allows the option to trade storage and computational cost - the adapted weights for a given user can be stored for faster classification. 

In our experiments with the GPDS-960 dataset, the proposed method obtains better results than WI systems in the literature, and approach the performance of WD systems, especially when few samples are available for training. With 5 reference signatures, the proposed method obtains 5.16\% EER (using a global threshold), compared to 9.05\% of a WI system and 5.25\% of a WD system. For a larger number of references the WD system still performs better, but the gap in performance is greatly reduced. Considering 12 reference signatures, the method obtains 4.39\% EER (with a global threshold), vs 3.74\% for the WD system, while being more scalable (single meta-classifier)
Our experiments transferring the meta-learner to other datasets show reduced performance, highlighting the need for better adaptation to new conditions, which will be explored in future work. Future work also includes considering a dynamic scenario, where the meta-classifier is updated as new training data is available. 


\bibliographystyle{IEEEtran}
\bibliography{biblio}

\begin{IEEEbiography}[{\includegraphics[width=1in,height=1.25in,clip,keepaspectratio]{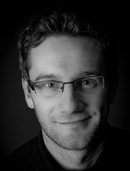}}]{Luiz G. Hafemann}
received his B.S. degree in Computer Science in 2008 and his M.Sc. degree in Informatics in 2014, both from the Federal University of Paraná, Curitiba, PR, Brazil. He received his Ph.D. degree in Systems Engineering in 2019 from the École de Technologie Supérieure, Université du Québec, in Montreal, QC, Canada. He is currently a researcher at Sportlogiq, applying computer vision models for sports analytics. His current interests include meta-learning, adversarial machine learning and group activity recognition.
\end{IEEEbiography}

\begin{IEEEbiography}[{\includegraphics[width=1in,height=1.25in,clip,keepaspectratio]{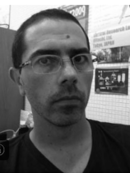}}]{Luiz S. Oliveira}
	received his B.S. degree in Computer Science from Unicenp, Curitiba, PR, Brazil, the M.Sc. degree in electrical engineering and industrial informatics from the Centro Federal de Educação Tecnológica do Paraná (CEFET-PR), Curitiba, PR, Brazil, and Ph.D. degree in Computer Science from École de Technologie Supérieure, Université du Québec in 1995, 1998 and 2003, respectively. From 2004 to 2009 he was professor of the Computer Science Department at Pontifical Catholic University of Paraná, Curitiba, PR, Brazil. In 2009, he joined the Federal University of Paraná, Curitiba, PR, Brazil, where he is professor of the Department of Informatics and head of the Graduate Program in  Computer Science. His current interests include Pattern Recognition, Machine Learning, Image Analysis, and Evolutionary Computation.
\end{IEEEbiography}

\begin{IEEEbiography}[{\includegraphics[width=1in,height=1.25in,clip,keepaspectratio]{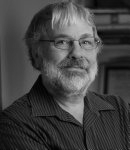}}]{Robert Sabourin}
joined the physics department of the Montreal University in 1977 where his main contribution was the design and the implementation of a microprocessor-based fine tracking system combined with a low-light level CCD detector. In 1983, he joined the staff of the École de Technologie Supérieure, Université du Québec, in Montreal, where he co-founded the Dept. of Automated Manufacturing Engineering where he is currently Full Professor and teaches Pattern Recognition, Evolutionary Algorithms, Neural Networks and Fuzzy Systems. In 1992, he joined also the Computer Science Department of the Pontificia Universidade Católica do Paraná (Curitiba, Brazil). Since 1996, he is a senior member of the Centre for Pattern Recognition and Machine Intelligence (CENPARMI, Concordia University). Since 2012, he is the Research Chair holder specializing in Adaptive Surveillance Systems in Dynamic Environments. Dr. Sabourin is the author (and co-author) of more than 450 scientific publications including journals and conference proceedings. His research interests are in the areas of adaptive biometric systems, adaptive classification systems in dynamic environments, dynamic classifier selection and evolutionary computation.
	
\end{IEEEbiography}

\end{document}